%% file: main_arxiv.tex
\documentclass[10pt,twocolumn,letterpaper]{article}

\usepackage[pagenumbers]{cvpr} % To force page numbers, e.g. for an arXiv version

\input{preamble}

\definecolor{cvprblue}{rgb}{0.21,0.49,0.74}
\usepackage{float}
\usepackage[pagebackref,breaklinks,colorlinks,citecolor=cvprblue]{hyperref}
\usepackage[all]{hypcap}
\usepackage{cuted}
\usepackage{capt-of}
\usepackage{multirow}
\usepackage{amsmath}
\usepackage{algorithm}
\usepackage{algpseudocode}
\usepackage[accsupp]{axessibility}  % Improves PDF readability for those with disabilities.

\usepackage[capitalize]{cleveref}
\crefname{section}{Sec.}{Secs.}
\Crefname{section}{Section}{Sections}
\Crefname{table}{Table}{Tables}
\crefname{table}{Tab.}{Tabs.}
\crefname{algorithm}{Alg.}{Algs.}
\Crefname{Algorithm}{Alg.}{Algs.}

\newcommand{\minisection}[1]{\vspace{2mm}\noindent{\textbf{#1}}}

\def\paperID{7110} % *** Enter the Paper ID here
\def\confName{CVPR}
\def\confYear{2024}
\def\modelName{SwiftBrush}

\title{\modelName\includegraphics[height=12.5pt, trim=-0.4cm 0.5cm -0.5cm 0cm]{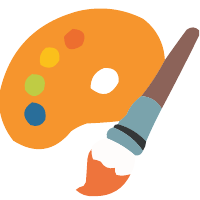}: One-Step Text-to-Image Diffusion Model with~Variational~Score~Distillation}

\author{
Thuan Hoang Nguyen \quad \quad \quad Anh Tran \\
VinAI Research, Vietnam \\
{\tt\small \{v.thuannh5,v.anhtt152\}@vinai.io}
}

\begin{document}

\makeatletter
\g@addto@macro\@maketitle{
  \begin{figure}[H]
  \setlength{\linewidth}{\textwidth}
  \setlength{\hsize}{\textwidth}
  \centering
  \includegraphics[trim=0 0.2cm 0 2cm, width=\textwidth]{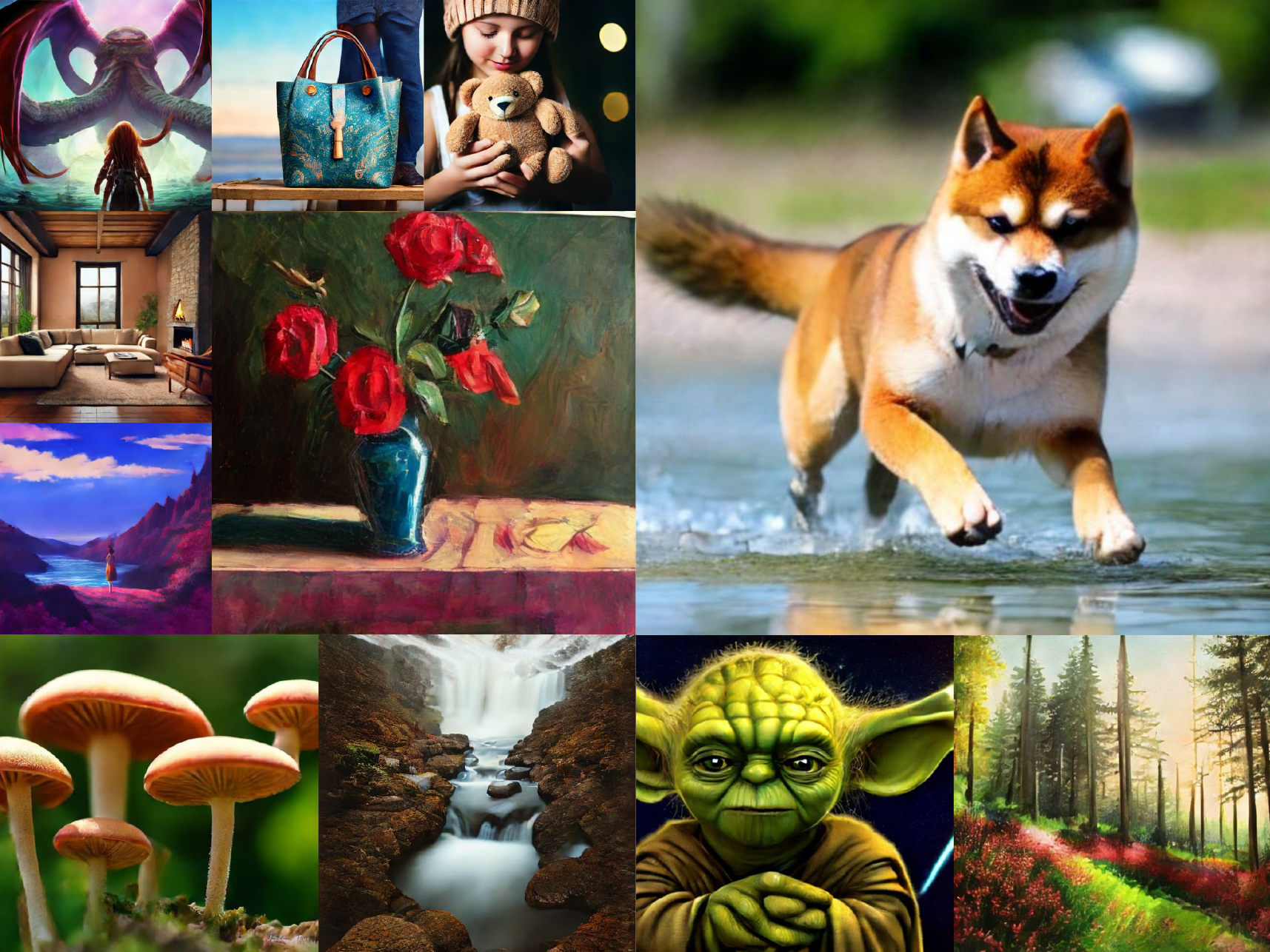}
  \vspace{-6mm}
  \caption{Samples generated in a \textbf{single step} by \modelName{}. Our distillation method can produce images of high-fidelity roughly \textbf{20 times} faster compared to Stable Diffusion.} \label{fig:teaser}
  \end{figure}
}
\makeatother

\maketitle

\input{sec/abstract}    
\input{sec/introduction}
\input{sec/related_work}
\input{sec/proposed_method}
\input{sec_arxiv/experiment}
\input{sec/conclusion}
{
    \small
    \bibliographystyle{ieeenat_fullname}
    \bibliography{main}
}

\input{sec/X_suppl}

\end{document}

%% file: preamble.tex
\usepackage[dvipsnames]{xcolor}

%% file: sec/abstract.tex
\begin{abstract}
Despite their ability to generate high-resolution and diverse images from text prompts, text-to-image diffusion models often suffer from slow iterative sampling processes. Model distillation is one of the most effective directions to accelerate these models. However, previous distillation methods fail to retain the generation quality while requiring a significant amount of images for training, either from real data or synthetically generated by the teacher model. In response to this limitation, we present a novel image-free distillation scheme named \textbf{\modelName}. Drawing inspiration from text-to-3D synthesis, in which a 3D neural radiance field that aligns with the input prompt can be obtained from a 2D text-to-image diffusion prior via a specialized loss without the use of any 3D data ground-truth, our approach re-purposes that same loss for distilling a pretrained multi-step text-to-image model to a student network that can generate high-fidelity images with just a single inference step. In spite of its simplicity, our model stands as one of the first one-step text-to-image generators that can produce images of comparable quality to Stable Diffusion without reliance on any training image data. Remarkably, \modelName{} achieves an FID score of \textbf{16.67} and a CLIP score of \textbf{0.29} on the COCO-30K benchmark, achieving competitive results or even substantially surpassing existing state-of-the-art distillation techniques.
\end{abstract}

%% file: sec/introduction.tex
\section{Introduction}\label{sec:intro}

%Various generative tasks, including image generation \cite{ho2020denoising, dickstein2015deep, nichol2021improving,song2021scorebased}, video synthesis \cite{ho2022video}, 3D modeling \cite{poole2023dreamfusion,watson2023novel,chan2023genvs}, audio generation \cite{kong2021diffwave}, and text creation \cite{gong2023diffuseq,li2022diffusionlm}, have come to rely on diffusion models as their standard tool. However, despite their impressive success, conventional diffusion models exhibit a noteworthy drawback: slow inference speed. The primary reason for this lies in the iterative nature of their sampling, which may present significant challenges when it comes to deployment on consumer devices.

%To overcome this drawback, researchers have proposed several methods to improve the sampling speed. One common approach involves accelerating the denoising process by enhancing ODE solvers \cite{lu2022dpmsolver,dockhorn2022genie}, which can generate images within 10-20 sampling steps. In another line of work, knowledge distillation techniques are used to improve the inference speed by training a student model that requires much fewer sampling steps than its teacher. In this work, we are interested in the latter approach for the task of text-to-image generation, where we train efficient models that only require one forward step to synthesize images from input captions.

Diffusion models are gaining significant attention from the research community as they have achieved remarkable results in various generative tasks, including image generation \cite{ho2020denoising, dickstein2015deep, nichol2021improving,song2021scorebased}, video synthesis \cite{ho2022video}, 3D modeling \cite{poole2023dreamfusion,watson2023novel,chan2023genvs}, audio generation \cite{kong2021diffwave}, and text creation \cite{gong2023diffuseq,li2022diffusionlm}. Especially, text-to-image diffusion models, which combine the power of language models and high-quality diffusion models \cite{SD,saharia2022photorealistic,ramesh2022hierarchical}, have revolutionized the way that people create visual content based on textual descriptions. Synthetic images indistinguishable from real photos are now possible with only a few clicks. Additionally, with techniques like ControlNet \cite{zhang2023adding} or DreamBooth \cite{ruiz2022dreambooth}, extra control to the generation outputs can be made to further boost the expressiveness of the text-to-image models and bring them closer to mass users. However, despite their impressive success, these models exhibit a noteworthy drawback of slow inference speed. The primary reason for this lies in the iterative nature of their sampling, presenting significant challenges for deployment on consumer devices.

%\anh{Several approaches have been proposed to improve the inference speed of diffusion-based text-to-image generation. However, the most effective direction is time-step distillation, which aims to reduce the number of sampling steps of the diffusion model without changing the network backbone and with minimum reduction in generation quality. Starting from the pioneering work \cite{salimans2022progressive}, more advanced and effective distillation methods have been proposed, and a few recent ones could train efficient student networks that only require one forward step to synthesize images from input captions.}

Several approaches have been proposed to improve the inference speed of diffusion-based text-to-image generation. However, the most effective direction is time-step distillation, which aims to reduce the number of sampling steps of the diffusion model with the network backbone unchanged and minimum reduction in generation quality. Starting from the pioneering work \cite{salimans2022progressive}, more advanced and effective distillation methods have been proposed, and a few recent ones could train efficient student networks that only require one forward step to synthesize images from input captions. In this work, we are interested in the distillation-based approach for one-step text-to-image generation. Particularly, \citet{meng2023on} employ a two-stage distillation 
% for conditional generative tasks like text-to-image
, where the student tries to first match its output with the classifier-free guided output of the teacher and then bootstrap itself to reduce the inference steps gradually. Such a method is complex and time-consuming, requiring at least $108$ A100 GPU days \cite{liu2023instaflow}. Based on the work of Song et al. \cite{song2023consistency}, Latent Consistency Model (LCM) \cite{luo2023latent} enforces the consistency of all points on the ODE trajectory in the latent space of Stable Diffusion (SD) \cite{SD} and unlike \cite{meng2023on}, they directly combine the two distillation stages into one. Although both Guided Distillation and LCM significantly reduce the number of inference steps to $2$ or $4$, their 1-step inference yields blurry and unsatisfying results. More sophisticated, Instaflow \cite{liu2023instaflow} proposes a technique called ``reflow'' to straighten the sampling trajectory of the teacher before distilling. While their proposed techniques proven to be superior to the original diffusion framework in one-step text-to-image distillation, InstaFlow requires an expensive 4-stage training schedule. Also, it is noteworthy that the effectiveness of these methods relies heavily on a large number of text-image pairs, which is not always the case due to the limited accessibility of such data.

In this work, we aim to develop a novel distillation method for one-step text-to-image generation with higher quality and a more approachable training process characterized by an image-free and straightforward mechanism. Building upon the recent advancements in text-to-3D techniques, our work takes inspiration from their ability to generate high-quality 3D Neural Radiance Fields (NeRF) without utilizing 3D ground truth data. Such a remarkable achievement is facilitated by employing a powerful pretrained 2D text-to-image model to assess whether the image rendered from NeRF is realistic or not, akin to the discriminator in GANs. This observation underpins the motivation for our SwiftBrush approach, as we recognize that these principles can be innovatively adapted for text-to-image generation making image supervision unnecessary. \modelName{}, therefore, emerges as a conceptually novel one-step text-to-image distillation process, uniquely capable of operating without 2D image supervision. This strategic adaptation not only opens up a change of view in the distillation of text-to-image diffusion models into a single-step generators but also points out the potential of blending principles from text-to-3D to text-to-image synthesis. 

As a result, our method has successfully enabled the generation of high-quality images with remarkable details. Quantitatively, our one-step model achieves promising zero-shot results, 16.6 for the FID score and 0.29 for the CLIP score on the MS COCO 2014 dataset (30K images). Notably, this is the first time a distilled one-step model outperforms the previous approaches without a single image used for training. A demonstration of generated images from our distilled model is shown in \cref{fig:teaser}.

%% file: sec/related_work.tex
\section{Related Work}\label{sec:related_work}

\minisection{Text-to-Image Generation.} Previously, text-to-image models relied on GANs and focused on only small-scale object-centric domains such as flowers and birds \cite{reed16generative,han2017stackgan}. Recently, these types of systems have advanced rapidly with the availability of web-scale datasets of text-image pairs such as LAION5B \cite{schuhmann2022laion5b}, large language models such as T5 \cite{raffel2020exploring}, and large vision-language models such as CLIP \cite{radford2021learning}. DALL-E \cite{ramesh2021zero-shot} was the first autoregressive model to demonstrate a remarkable zero-shot ability in creating images from text captions by simply scaling up both the network and dataset size. Following that, a plethora of methods have been used for text-to-image generation, including masked generative transformers \cite{chang2023muse}, GAN-based, and diffusion models. Based on Latent Diffusion Model \cite{rombach2022highresolution}, Stable Diffusion is an open-source diffusion-based generator that has gained widespread popularity among artists as well as researchers. Recently, text-to-image GANs are receiving attention again with notable work like StyleGAN-T \cite{sauer2023stylegant}, or GigaGAN \cite{kang2023scaling}. While they are extremely fast as no iterative sampling is required, they lacks behind diffusion models in terms of scalability and realism due to the fact that they suffer from instability and the infamous issue ``mode collapse''. To the best of our knowledge, so far only two above-mentioned work succeeded in applying GANs for text-to-image using tons of intricate techniques and auxiliary modules. Conversely, diffusion models can be easily scale up to generate high-quality and sharp samples, yet they require lengthy sampling process. 
% \anh{I would talk lightly about GAN-based and talk more about diffusion-based approaches. Discuss the pros and cons of each (particularly the weakness of GAN, otherwise no one needs diffusion models), and our method has both advantages}. 
Here, our objective is to create a large-scale generative model that combines the speed of GANs with the quality of diffusion models.

\minisection{Text-to-3D Generation.} Instead of learning a generative model directly from 3D supervision, many research works have utilized the rich knowledge of 2D prior for 3D generation tasks. For instance, one of the earliest works, Dream Fields \cite{jain2021zeroshot}, leverages CLIP \cite{radford2021learning} to guide the generated images so that rendered images from many camera views highly align with the text caption. On the other hand, pioneer works like DreamFusion \cite{poole2023dreamfusion} and Score Jacobian Chaining \cite{wang2023score} (SJC) propose two different but equivalent frameworks where a 2D text-to-image generative prior is used to generate 3D objects from textual descriptions. Subsequent works seek to improve these text-to-3D methods in various ways. Among them, Magic3D \cite{lin2023magic3d} and ProlificDreamer \cite{wang2023prolificdreamer} are two remarkable works that effectively enhance the generation quality of 3D assets. Magic3D consists of two stages where they first obtain a low-resolution 3D representation via a DreamFusion-style optimization and then refine its texture to get the final high-fidelity 3D mesh. Meanwhile, ProlificDreamer proposes Variational Score Distillation (VSD) involving a secondary teacher to bridge the gap between the teacher and the 3D NeRF. Influenced by the latter work, our method adapts the same technique for one-step text-to-image generation. Similarly, Score GAN \cite{franceschi2023unifying} and Diff-Instruct \cite{luo2023diffinstruct} use a VSD-like loss to distill a pretrained diffusion model to a one-step generator.

\minisection{Knowledge Distillation} introduced by  \citet{hinton2014distilling} is one of the methods falling under the umbrella of transfer learning. This algorithm draws its main inspiration from the human learning process, where knowledge is conveyed from a more knowledgeable teacher to a less knowledgeable student. In the context of diffusion models, works have been proposed for faster sampling of student models from a pre-trained diffusion teacher. The most straightforward approach involves direct distillation \cite{luhman2021knowledge}, wherein a student model is supervised from the teacher's output after sampling for 50-1000 steps, which can be prohibitive. Recent work \cite{song2023consistency, salimans2022progressive} avoids the lengthy 
sampling during distillation by various bootstrapping methods. Afterward, a series of methods \cite{luo2023latent, liu2023instaflow, meng2023on} adapt these methods into text-to-image settings and inherit their limitation of being image-dependent in training. In contrast, a concurrent work \cite{gu2023boot} learns a time-conditioned model to predict the outcome of the teacher at any given time step, all without the need for image supervision. Likewise, our approach is similarly image-free yet achieves significantly better results and offers a much simpler distillation design.

%% file: sec/proposed_method.tex
\section{Proposed Method}\label{sec:proposed_method}

\subsection{Preliminary}\label{sec:preliminary}
\minisection{Diffusion Models} consist of two processes: a forward process that gradually adds noise and another reverse process to predict the distribution of gradually denoised data points. The forward process incrementally introduces noise to an initial data point $x_0$, drawn from the distribution $q_0(x_0)$, resulting in $q_t(x_t \vert x_0) = \mathcal{N}(\alpha_t x_0, \sigma_t^2 I)$, where $\{(\alpha_t, \sigma_t) \}_{t=1}^T$ is the noise schedule. The data point $x_t$ at each time-step $t$ is drawn from $q_t(x_t \vert x_0)$ using a noise $\epsilon \sim \mathcal{N}(0, I)$:
\begin{equation}
    x_t = \alpha_t x_0 + \sigma_t \epsilon. \label{eq:xt}
\end{equation}
After $T$ time steps, the final data point is expected to be pure Gaussian noise $x_T \sim \mathcal{N}(0, I)$. 

Conversely, the backward process starts from $x_T \sim \mathcal{N}(0, I)$ and iteratively denoises for $T$ steps where at each step $t$, noisy variable $x_{t+1}$ is converted to less-noisy variable $x_t$, guided by a model $\epsilon_\psi$ that predicts the noise $\epsilon$ in \cref{eq:xt}. The network weights $\psi$ are trained by minimizing:
\begin{equation}
    \mathcal{L}_{uncond}(\psi) = \mathbb{E}_{t, \epsilon \in \mathcal{N}(0,1)} \Vert \epsilon_\psi(x_{t}, t) - \epsilon \Vert_2^2
    \label{eq:LossUncondDiff}
\end{equation}
where $t$ is uniformly samples within $\{1, \dots, T\}$.

While common diffusion models operate in the pixel space, Latent Diffusion Models (LDMs)\cite{rombach2022highresolution}  model the diffusion process in the latent space of a pre-trained regularized autoencoder, which features representations of smaller size, thus enhancing both training and inference efficiency.

% \minisection{Latent Diffusion Models (LDMs)} aim to enhance both training and inference efficiency in diffusion models, which traditionally operate directly within the pixel-space. They achieve this by modeling the diffusion process in the latent space generated by a pre-trained regularized autoencoder. This latent space typically features representations of smaller resolution compared to the original pixel-space. As an alternative to cascaded diffusion methods that use one or more super-resolution models to upscale a low-resolution image to a higher target resolution, LDMs offer a more efficient and streamlined approach.

\minisection{Text-to-Image Diffusion Models.} In contrast to unconditional diffusion models that generate outputs freely, text-conditioned diffusion models steer the sampling process using an extra prompt $y$. This guides the model to produce outputs that are not only photorealistic but also closely adhere to the provided text description. The objective to train such models is formulated as follows:
\begin{equation}
    \mathcal{L}_{diff}(\psi) = \mathbb{E}_{t, y, \epsilon \in \mathcal{N}(0,1)} \Vert \epsilon_\psi(x_{t}, t, y) - \epsilon \Vert_2^2
    \label{eq:LossConddiff}
\end{equation}
which is slightly different from the unconditional diffusion loss \cref{eq:LossUncondDiff}. Due to the prompt conditioning, the model can deliver more controllable generation compared to its unconditional counterparts. Nevertheless, the implementations of many leading methods \cite{saharia2022photorealistic, ramesh2022hierarchical, balaji2022ediffi} remain inaccessible to the public. Stable Diffusion, which primarily utilizes the LDM framework, stands out as the first openly available large-scale model, significantly propelling text-to-image synthesis's widespread adoption and versatility.

\minisection{Score Distillation Sampling (SDS)} is a distillation technique effectively applied in generating 3D assets \cite{poole2023dreamfusion,wang2023score,lin2023magic3d}. It utilizes a pre-trained text-to-image diffusion model, which predicts diffusion noise from text condition $y$, denoted as $\epsilon_{\psi}(x_t, t, y)$. The method optimizes a single 3D NeRF, parameterized by $\theta$, to align with a given text prompt. Given camera parameters $c$, a differentiable rendering function $g(\cdot, c)$ is used to render an image at the camera view $c$ from the 3D NeRF. Here, the rendered image $g(\theta, c)$ is utilized to optimize the weight $\theta$ through a loss function whose gradient can be approximated by:
\begin{equation}
\nabla_\theta \mathcal{L}_{SDS} = \mathbb{E}_{t, \epsilon, c} \left[w(t)(\epsilon_{\psi}(x_t, t, y) - \epsilon)\frac{\partial g(\theta, c)}{\partial \theta}\right]
\end{equation}
where $t \sim \mathcal{U}(0.02T, 0.98T)$, $T$ is the maximum timesteps of the diffusion model, $\epsilon \sim \mathcal{N}(0, I)$, $x_t = \alpha_t g(\theta, c) + \sigma_t \epsilon$, $y$ is the input text, and $w(t)$ is a weighting function

Despite the advancements in text-to-3D synthesis, empirical studies \cite{poole2023dreamfusion,wang2023score} indicate that SDS often encounters issues like over-saturation, over-smoothing, and reduced diversity. The same degradations can be observed if we naively apply SDS in our framework, as shown in Sec. \ref{sec:analysis}.

\minisection{Variational Score Distillation (VSD)} is introduced in ProlificDreamer \cite{wang2023prolificdreamer} to address previously mentioned issues of SDS by modifying the loss slightly:
\begin{equation}\label{eq:loss_vsd}
\begin{split}
\nabla_\theta \mathcal{L}_{VSD} = \mathbb{E}_{t, \epsilon, c} [&w(t)(\epsilon_{\psi}(x_t, t, y) \\
& - \epsilon_{\phi}(x_t, t, y, c))\frac{\partial g(\theta, c)}{\partial \theta}]
\end{split}
\end{equation}
Again, $x_t = \alpha_t g(\theta, c) + \sigma_t \epsilon$ is the noisy observation of the rendered image at camera view $c$. VSD sets itself apart from SDS by introducing an additional score function tailored to the images rendered from the 3D NeRF at the camera pose $c$. This score is derived by fine-tuning a diffusion model $\epsilon_{\phi}(x_{t}, t, y, c)$ with the diffusion loss described below:
\begin{equation}\label{eq:loss_lora}
    \min_{\epsilon_{\phi}} \mathbb{E}_{t, c, \epsilon} \Vert \epsilon_{\phi}(x_{t}, t, y, c) - \epsilon \Vert_2^2
\end{equation}
As proposed in ProlificDreamer, $\epsilon_{\phi}$ is parameterized by Low-Rank Adaption \cite{hu2022lora} (LoRA) and initialized from the pre-trained diffusion model $\epsilon_{\psi}$ with some added layers for conditioning the camera view $c$. Note that at each iteration $i$ of the optimization process, $\epsilon_\phi$ needs to adapt with the current distribution of $\theta$. Therefore, ProlificDreamer interleaves between finetuning $\epsilon_\phi$ and optimizing $\theta$. With these algorithmic enhancements, ProlificDreamer has significantly advanced its capabilities, enabling the generation of NeRFs and the creation of exceptional textured meshes. Such improvement directly inspires us to adapt VSD to the task of one-step text-to-image diffusion-based distillation.

\subsection{\modelName{}}\label{sec:main_method}

\minisection{Motivation.} While SDS and VSD are designed explicitly for the text-to-3D generation task, they loosely connect to that goal via the rendered image $g(\theta, c)$ of the 3D NeRF. As a matter of fact, instead of using NeRF rendering, we can replace it with any function that outputs 2D images to suit our needs. Inspired by this motivation, we propose to substitute the NeRF rendering $g(\theta, c)$ with a text-to-image generator that can directly synthesize a text-guided image in one step, effectively converting the text-to-3D generation training into one-step diffusion model distillation.

\begin{figure*}
    \centering
    \includegraphics[width=.94\textwidth]{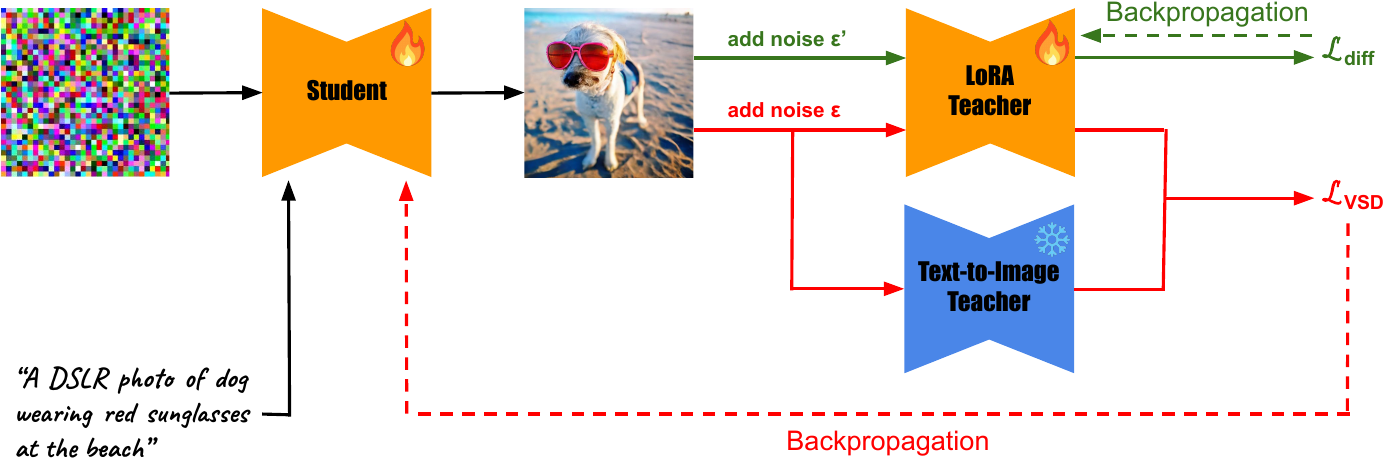}
    \vspace{-2mm}
    \caption{\textbf{\modelName{} overview.} Our system allows training a one-step text-to-image student network from a frozen pretrained teacher with the help of an additional trainable LORA teacher. The student network takes inputs a text prompt and random noise. The output of the student is then added noise and sent to two teachers, alongside the prompt and a randomly drawn timestep, to compute the gradient of VSD loss, which backpropagates back to the student. Apart from that, the LoRA teacher is also updated using diffusion loss. Similar to ProlificDreamer \cite{wang2023prolificdreamer}, we update the LoRA teacher (\color{ForestGreen}green flow\color{black}) and the student (\color{red}red flow\color{black}) alternatively.}
    \label{fig:system_figure}
    \vspace{-2mm}
\end{figure*}

\minisection{Design Space.} We adopt the same approach as ProlificDreamer \cite{wang2023prolificdreamer}, with modifications to the design space to suit our task better. To begin with, we make use of two teacher models: a pretrained text-to-image teacher $\epsilon_\psi$ and one additional LoRA teacher $\epsilon_\phi$. Also, we remove the conditioning on camera view $c$ from the LoRA teacher as it is unnecessary in our case, and we use classifier-free guidance for both teachers. Then, we replace the NeRF, which is overfitted for a specific user-provided prompt in a text-to-3D setting, with a generalized one-step text-to-image student model $f_{\theta}(z, y)$. Our student  model $f_{\theta}$ takes a random Gaussian noise $z$ and a text prompt $y$ as inputs. Both the LoRA teacher and student model are initialized with the weight of the text-to-image teacher. Next, we train both the student model and the LoRA teacher alternately using \cref{eq:loss_vsd} and \cref{eq:loss_lora} while freezing the text-to-image teacher. A pseudo-code and system figure can be viewed in \cref{alg:pseudocode} and \cref{fig:system_figure}. %, respectively.

\minisection{Student Parameterization.} Given a pretrained text-to-image diffusion model $\epsilon_{\theta}$, it is possible to directly use its output for the student model, i.e., $f_\theta(z, y) = \epsilon_{\theta}(z, T, y)$, where $T$ is the maximum timestep of the pretrained model. However, in our case, the pretrained model of choice is Stable Diffusion, which is inherently designed to predict the added noise $\epsilon$. In contrast, our objective is to refine the student model so that it predicts a clean and noise-free $x_0$. Therefore, such a naive method results in a large domain gap between what we want the student to learn and the output of the student. For the sake of easy training, we empirically re-parameterize the student output as follows: 
$$f_\theta(z, y) = \frac{z - \sigma_T\epsilon_{\theta}(z, T, y)} {\alpha_T}$$ 
which is a realization of \cref{eq:xt} if we set $t = T$, $x_t = z$, $\epsilon \approx \epsilon_{\theta}(z, T, y)$ and $x_0 \approx f_\theta(z, y)$. So basically, this re-parameterization transforms the noise-prediction output of the pretrained model into the ``predicted $x_0$'' form, which is empirically demonstrated in \cref{sec:analysis} to be helpful for the student model to learn.

\begin{algorithm}[t]
\caption{\modelName{} Distillation}
\begin{algorithmic}[1]
\State \textbf{Require}: a pretrained text-to-image teacher $\epsilon_\psi$, a LoRA teacher $\epsilon_\phi$, a student model $f_\theta$, two learning rates $\eta_1$ and $\eta_2$, a weighting function $\omega$, a prompts dataset $Y$, the maximum number of time steps $T$ and the noise schedule $\{(\alpha_t, \sigma_t) \}_{t=1}^T$ of the teacher model
\State \textbf{Initialize:} $\phi \gets \psi$, $\theta \gets \psi$
\While{not converged}
    \State Sample input noise $z \sim \mathcal{N}(0, I)$
    \State Sample text caption input $y \sim Y$
    \State Compute student output $\hat{x}_0 = f_\theta(z, y)$
    \State Sample timestep $t \sim \mathcal{U}(0.02T, 0.98T)$
    \State Sample added noise $\epsilon \sim \mathcal{N}(0, I)$
    \State Compute noisy sample $\hat{x}_t = \alpha_t \hat{x}_0 + \sigma_t \epsilon$
    \State $\theta \gets \theta - \eta_1 \left[ \omega(t) \left( \epsilon_{\psi}(\hat{x}_t, t, y) - \epsilon_\phi(\hat{x}_t, t, y) \right) \frac{\partial \hat{x}_0}{\partial \theta} \right]$
    \State Sample timestep $t' \sim \mathcal{U}(0, T)$
    \State Sample added noise $\epsilon' \sim \mathcal{N}(0, I)$
    \State Compute noisy sample $\hat{x}_{t'} = \alpha_{t'} \hat{x}_0 + \sigma_{t'} \epsilon'$
    \State $\phi \gets \phi - \eta_2 \nabla_\phi \| \epsilon_\phi(\hat{x}_{t'}, t', y) - \epsilon' \|^2$
\EndWhile
\State \Return trained student model $f_\theta$
\end{algorithmic}
\label{alg:pseudocode}
\end{algorithm}

%% file: sec_arxiv/experiment.tex
\section{Experiments}\label{sec:experiment}

\subsection{Experimental Setup}\label{sec:setup}
\minisection{Evaluation Metrics.} We conduct our evaluation process by assessing the performance of \modelName{} against several well-established methods, such as Guided Distillation \cite{meng2023on}, Instaflow \cite{liu2023instaflow}, LCM \cite{luo2023latent}, and BOOT \cite{gu2023boot}, across two zero-shot text-to-image benchmarks: COCO 2014 \cite{lin2014microsoft} and Human Preference Score v2 \cite{wu2023human} (HPSv2). On the COCO 2014, we follow the conventional evaluation protocol \cite{sauer2023stylegant, kang2023scaling, saharia2022photorealistic, rombach2022highresolution}, computing FID and CLIP score on a 30K subset of its validation set as our primary metrics. Here, we use ViT-G/14 as the backbone for evaluating CLIP scores. In the case of HPSv2, we adopt their evaluation procedure to assess text-image alignment across four different aspects: anime, concept-art, paintings, and photos. 

\minisection{Experiment Protocols.} In these experiments, we rely on Stable Diffusion v2.1 \cite{SD} for our two teacher models, and whenever possible, we directly utilize the reported values from other research. As for InstaFlow and LCM, we use the their provided pretrained models to generate images for evaluation. Conversely, we are compelled to re-implement BOOT since they did not provide any metrics for their text-to-image models, and their codebase remains entirely closed.

\minisection{Implementation Details.} Similar to prior research, we use student models with the same architectures as those of the teachers and initialize all of the student parameters with the teacher model. During training, a guidance scale of $4.5$ is used for both teachers. For the LoRA teacher model, we apply a learning rate of 1e-3, a LoRA rank of 64, and an alpha value of 108. Meanwhile, the student model's learning rate is set at 1e-6 with exponential moving average (EMA) used every iteration. Adam optimizer \cite{kingma2017adam} and a batch size of 64 (16 per GPU) are used to train both the student and LoRA teacher for around 65K iterations. Thanks to the image-free nature of \modelName{}, we do not require access to any image training set, and only the prompts are needed to distill the generator. Here, we utilize 4M captions from the large-scale text-image dataset JourneyDB \cite{pan2023journeydb}, which we subsequently deduplicated to 1.38M. Regarding BOOT, we closely follow their implementation guidelines to the best of our abilities and train the student model until its convergence, except that we use Stable Diffusion v2.1 instead of DeepFloyd IF \cite{IF} as the teacher. Here, its blurry results (\cref{fig:qualitative}) compared to the original work could be due to the different teacher.

\subsection{Results}\label{sec:results}

\begin{figure}[t]
    \begin{center}
        \includegraphics[width=0.45\textwidth]{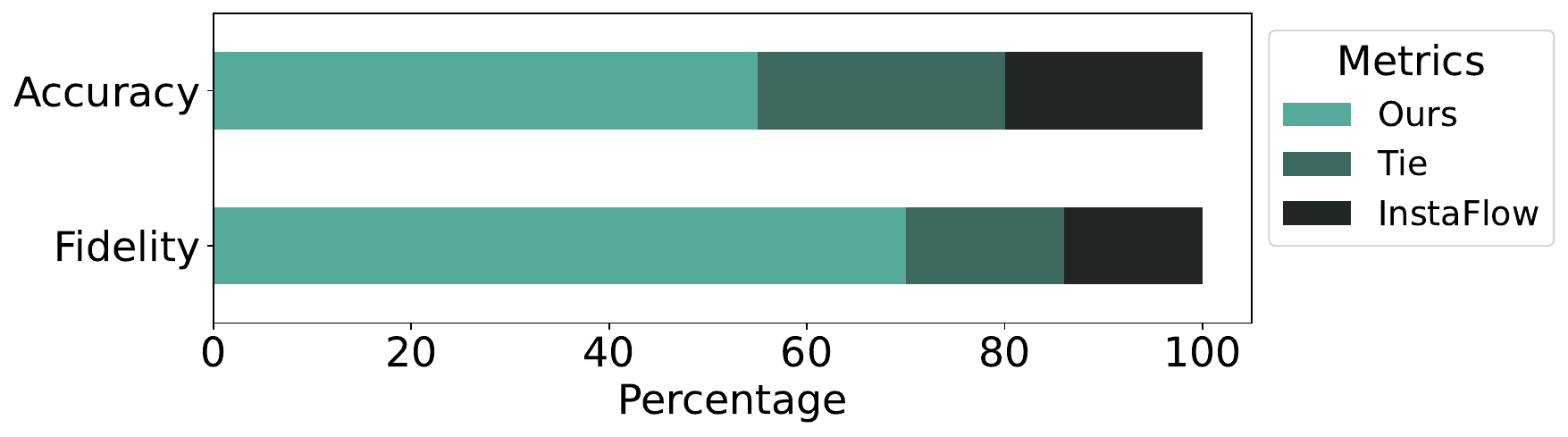}
    \end{center}
    \vskip -0.28in
    \caption{\textbf{User Study.} We ask 400 users to compare the generated images of InstaFlow and our method using 30 random text prompts. We report the rate at which each method is selected as better for image quality and text-to-image accuracy.}
    \label{fig:survey}
    \vskip -0.15in
\end{figure}

\minisection{Quantitative Results.} We first evaluate the proposed method on standard text-to-image generation benchmarks. The quantitative comparison with the previous distillation, both image-dependent and image-free, and the original Stable Diffusion teacher is shown in \cref{tab:fid_clip}. Despite being trained with only text captions, \modelName{} significantly improves the quality of 1-step inference and achieves better photo-realism as well as text-image alignment against other methods, with exceptions of InstaFlow and Stable Diffusion (25 sampling steps) in FID score. However, it is worth noting that Instaflow is so complex, requiring a 4-stage distillation process, while ours is end-to-end and straightforward. We further conduct a quantitative evaluation with the HPSv2 score, a metric trained to capture human preference. Again, we obtain the best score in all aspects among all mentioned works, which is only a slight degradation compared to the original Stable Diffusion teacher (\cref{tab:hpsv2}). Moreover, according to the anonymous survey carried out by us (\cref{fig:survey}), our method outperforms Instaflow in terms of image fidelity and text-to-image accuracy.

\begin{figure*}[t]
    \small
    \begin{center}
        \includegraphics[width=\textwidth]{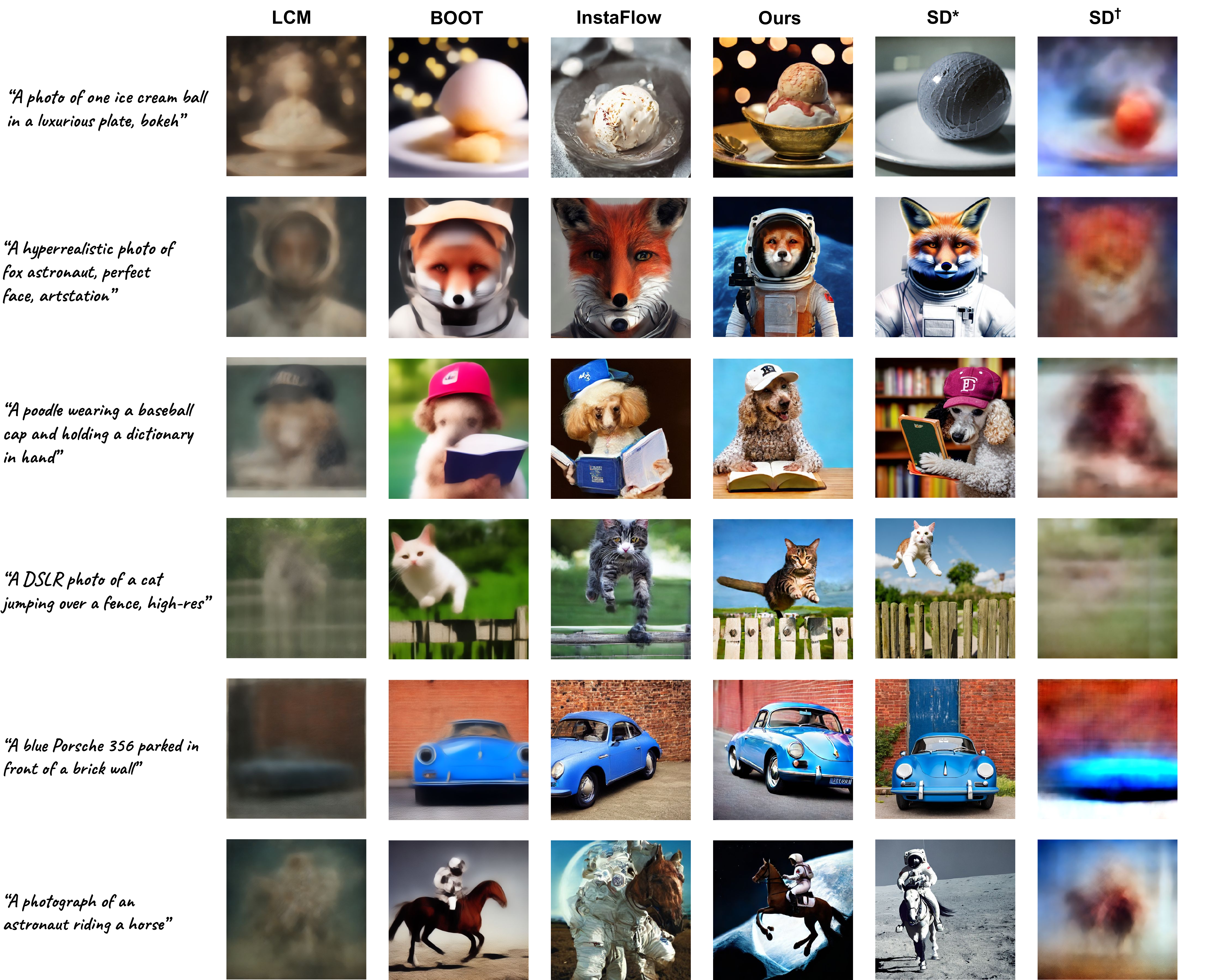}
    \end{center}
    \vskip -0.22in
    \caption{Sample images generated by LCM, BOOT, Instaflow, Ours, SD with 25 sampling steps (SD$^*$) and SD with 1 sampling step (SD$^\dag$). Images at the same row are sampled from the same text prompt, while images at the same column are from the same model. We use DPM-Solver \cite{lu2022dpmsolver} with guidance scale $7.5$ to sample for Stable Diffusion}
    \label{fig:qualitative}
    \vskip -0.15in
\end{figure*}

\minisection{Qualitative Results.} We report the visual comparison between our method and the mentioned baseline methods for the one-step text-to-image generation task. As seen in \cref{fig:qualitative}, LCM outputs blurry and over-smoothed images in the 1-step regime. BOOT faithfully produces images that match well with the text caption, but its results still fall short in terms of fidelity. In contrast, a much better quality can be observed in the image produced by Instaflow and \modelName{}. InstaFlow, nevertheless, cannot handle complex or unusual input prompts properly, which can be noticed in the third and fifth images at the third column of \cref{fig:qualitative}. On top of that, it occasionally produces non-realistic results even for simple prompts like in \cref{fig:qualitative} at the fourth image of the third column. When, in fact, the outputs generated by \modelName{} are not only of exceptional quality but also exhibit a remarkable alignment between text descriptions and the corresponding images. Moreover, our results are comparable with those sampled from Stable Diffusion with 25 steps; whereas it is understandable that with only one step, Stable Diffusion cannot create outputs of good quality. 

\begin{table}[t]
\centering
\begin{tabular}{l|c|c|c}
\hline
Method & Steps & FID-30K $\downarrow$ & CLIP-30K $\uparrow$ \\
\hline
Guided Distillation$^\dag$ & 1 & 37.3 & 0.27 \\
LCM$^\dag$ & 1 & 35.56 & 0.24 \\
Instaflow & 1 & \textbf{13.10}$^\dag$ & \underline{0.28}$^\S$ \\
BOOT$^\ddag$ & 1 & 48.20 & 0.26 \\
Ours & 1 & \underline{16.67} & \textbf{0.29} \\
\hline
%SD 2.1 & 50 & 15.64 & 0.24 \\
SD 2.1$^*$ & 25 & 13.45 & 0.30 \\
SD 2.1$^*$ & 1 & 202.14 & 0.08 \\
\bottomrule
\end{tabular}
\vspace{-2mm}
\caption{Comparison of our method against other works based on FID  metric and CLIP score on the COCO 2014 dataset. $^\dag$ means that we obtain the numbers from the corresponding papers. $^\S$ means that we obtain the numbers using the provided pretrained models of the corresponding papers. $^\ddag$ means that we re-implement the work and report the numbers. $^*$ means that we report the number using the pretrained models. We use DPM-Solver \cite{lu2022dpmsolver} with guidance scale $7.5$ to sample for SD. \textbf{Bold} and \underline{underlined} numbers are best and second best for one-step models.}
\label{tab:fid_clip}
\vspace{-2mm}
\end{table}

\begin{table}[t]
\centering
\begin{tabular}{l|cccc}
\hline
\multirow{2}{*}{Models} & \multicolumn{4}{c}{Human Preference Score v2 $\uparrow$}                                                                           \\ \cline{2-5} 
                        & \multicolumn{1}{l}{Anime} & \multicolumn{1}{l}{Photo} & \multicolumn{1}{l}{Concept Art} & \multicolumn{1}{l}{Paintings} \\ \hline
LCM$^\dag$                     & 22.61                     & 22.71                     & 22.74                           & 22.91                         \\
InstaFlow$^\dag$               & \underline{25.98}                     & \underline{26.32}                     & \underline{25.79}                           & \underline{25.93}                         \\
BOOT$^\ddag$                    & 25.29                     & 25.16                     & 24.40                           & 24.61                         \\
Ours                    & \textbf{26.91}                     & \textbf{27.21}                     & \textbf{26.32}                           & \textbf{26.37}                         \\ \hline
SD 2.1$^*$                  & 27.48                     & 26.89                     & 26.86                           & 27.46     \\ 
\hline
\end{tabular}
\vspace{-2mm}
\caption{Comparison of our method against other works based on HPSv2 score in 1-step regime. $^\dag$ means that we obtain the score using the provided pretrained models of the corresponding papers. $^\ddag$ means that we re-implement the work and report the score by ourselves. $^*$ means that we obtain the score from \cite{wu2023human}. \textbf{Bold} and \underline{underlined} numbers are best and second best, respectively.}
\label{tab:hpsv2}
\vspace{-4.5mm}
\end{table}

\subsection{Analysis}\label{sec:analysis}

\begin{figure*}[t]
\centering
\begin{subfigure}{.5\textwidth}
  \centering
  \includegraphics[width=.8\linewidth]{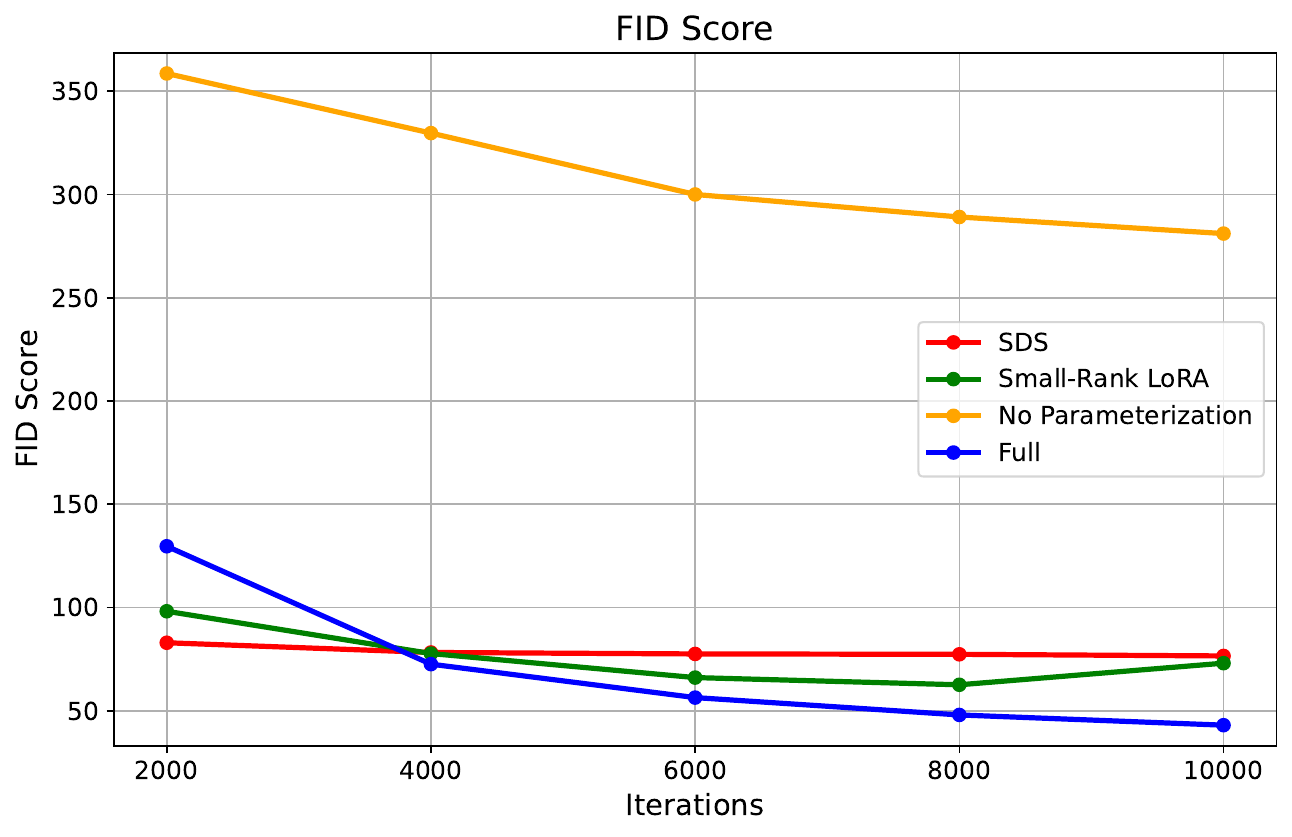}
  \caption{FID curves of four configurations}
  \label{fig:sub1}
\end{subfigure}%
\begin{subfigure}{.5\textwidth}
  \centering
  \includegraphics[width=.8\linewidth]{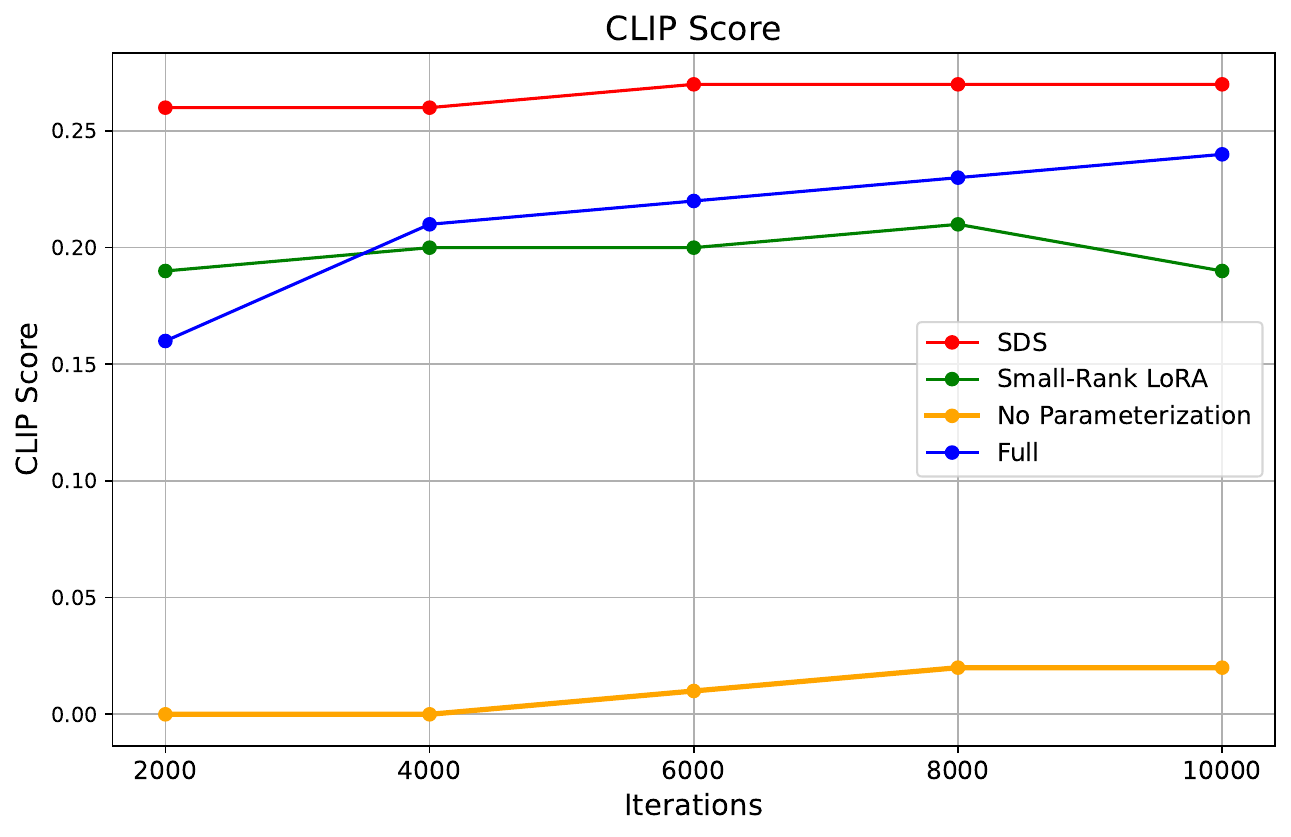}
  \caption{CLIP curves of four configurations}
  \label{fig:sub2}
\end{subfigure}
\vspace{-3mm}
\caption{Ablation study on the impact of the LoRA teacher as well as the student parameterization. We train all four configurations with same hyperparameters described in \cref{sec:experiment} but only with 10K iterations to observe their training dynamics at the early phase. Then, we evaluate FID and CLIP score on COCO 2014 but with only a 5K subset of images. Here, Full means all of our proposed techniques are used. No Parameterization means we exclude the re-parameterization trick. Small-Rank LoRA means we use only 4-rank LoRA teacher. SDS means we remove the LoRA teacher and use only SDS loss.}
\vspace{-3mm}
\label{fig:fid_clip_coco5k}
\end{figure*}

\begin{figure}[t]
    \begin{center}
        \includegraphics[width=.47\textwidth]{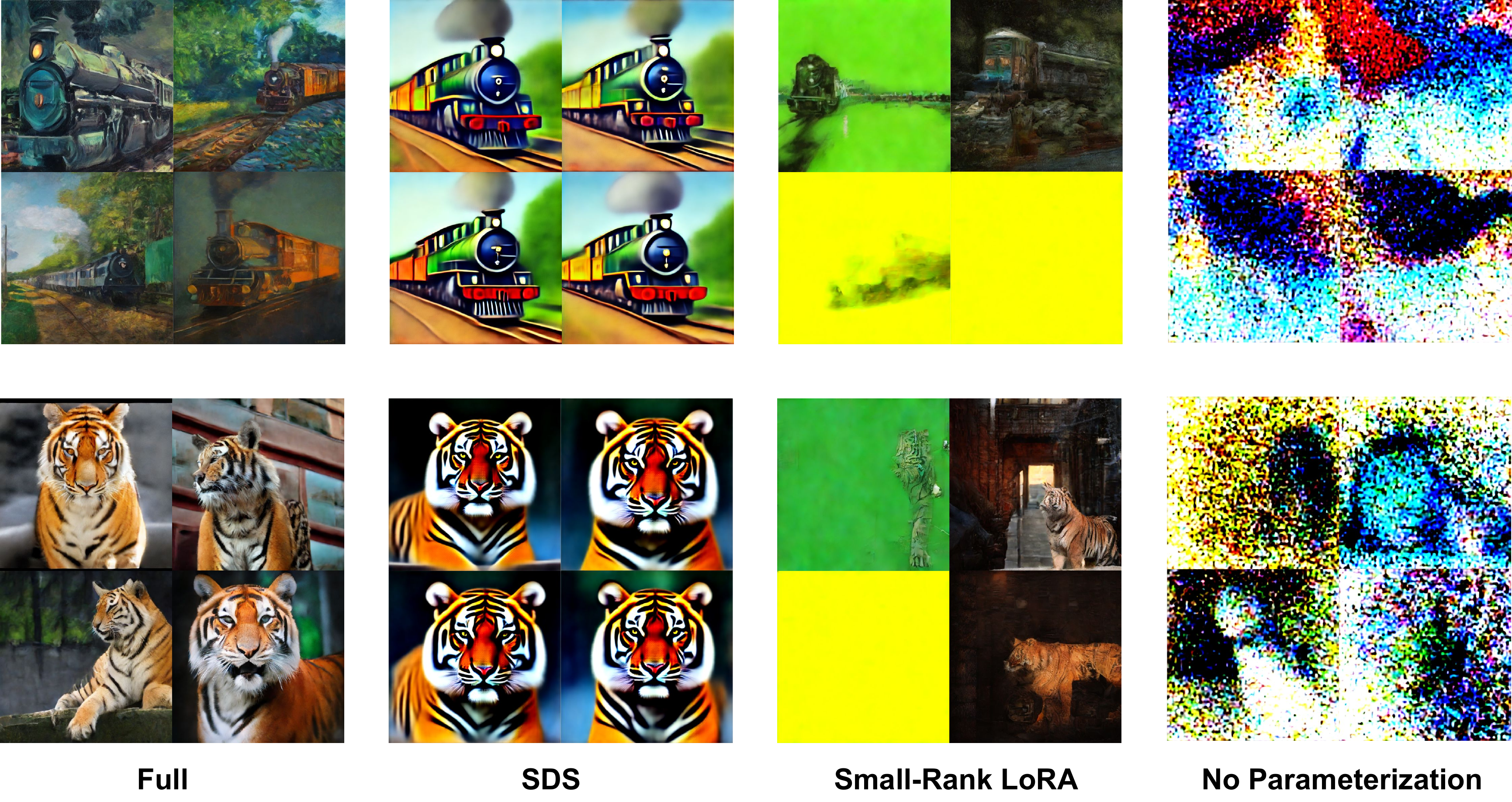}
    \end{center}
    \vskip -0.22in
    \caption{Visual results of the ablative study, where the first two rows are generated with the input prompt \textit{``An oil painting of a train``}, whereas the remaining two are sampled from \textit{``An DSLR photo of a tiger in the city``}.}
    \label{fig:visual_ablate}
    \vskip -0.15in
\end{figure}

\minisection{The importance of LoRA teacher.} In \cref{fig:visual_ablate}, the importance of including the LoRA teacher in the training process is clearly highlighted. We compare the student outputs sampled with different input noises after 10K iteration when using the same text prompts. With the guidance of the LoRA teacher, the student progressively learns to create realistic images ($1^{st}$ column). In contrast, in the absence of an effective LoRA teacher, the student's early training outputs, despite being highly aligned with the text prompt, are overly saturated and unrealistic ($2^{nd}$ column). These results indicate a fundamental breakdown in the student model's ability to mimic the teacher model's distribution, as can be further demonstrated based on the flat red curve in both FID and CLIP score in \cref{fig:fid_clip_coco5k} (a) and (b).

In addition, even when employing the LoRA teacher, it is crucial to use a sufficiently large rank in the LoRA weight for stable distillation of the student model. Without this, the model tends to undergo severe mode collapse despite initially producing reasonably good results. The green curves in \cref{fig:fid_clip_coco5k} clearly illustrate that, while the FID score of a student model with a 4-rank LoRA weight initially decreases, it deteriorates starting from 8K steps, resulting in lousy images in \cref{fig:visual_ablate} ($3^{rd}$ column). Additionally, the corresponding CLIP score behaves similarly where there is a slight rise before it worsens in the end. Besides, we also provide extra visuals for this ablative study in \cref{sec:additional_qualitative}.

\minisection{The importance of student parameterization.} Apart from the LoRA teacher, the re-parameterization of the student (\cref{sec:proposed_method}) also plays an important role in the \modelName{} training. Without it, the student is unable to to create realistic images and it only produces noisy results, as can be noticed from the $4^{th}$ column of \cref{fig:visual_ablate}. This can be further backed up by the orange curve in both FID and CLIP score in \cref{fig:fid_clip_coco5k} (a) and (b), where student without re-parameterization perform consistently worst among all configurations.

\begin{figure}[t]
    \small
    \begin{center}
    \includegraphics[width=.47\textwidth]{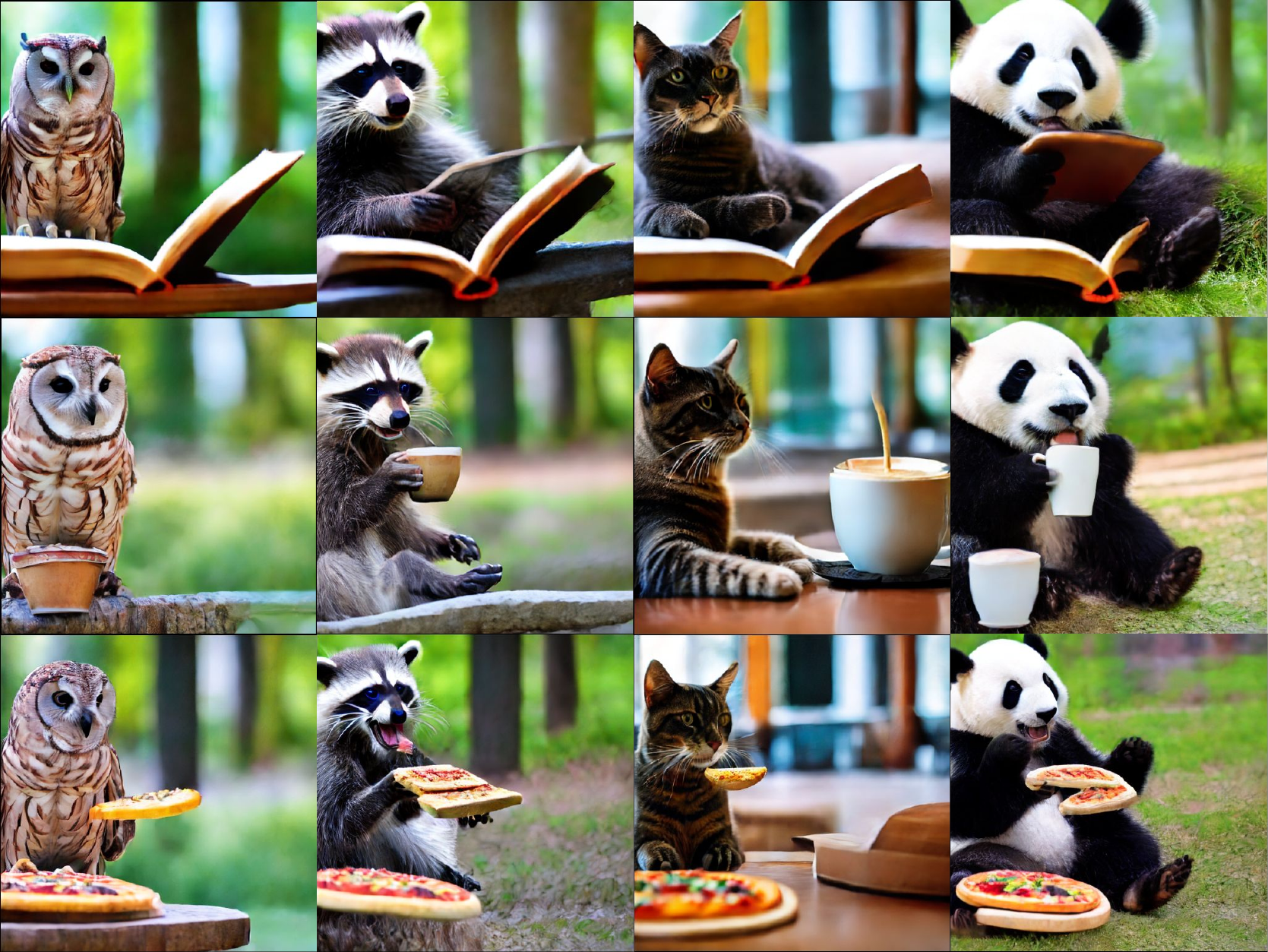}
    \end{center}
    \vskip -0.22in
    \caption{Results of interpolating the input prompts. The prompts used here are selected from a standard template: `A DSLR photo of a \{animal\} \{action\}'. Here, `animal' is one of following: owl, raccoon, cat, or panda, while `action' is one of following: reading a book, drinking a latte or eating a pizza.}
    \label{fig:interpolate}
    \vskip -0.15in
\end{figure}

\minisection{Interpolating Latent.} We conduct a controlled experiment where the noise input $z$ remains the same while varying the input prompt $y$ (\cref{fig:interpolate}). Interestingly, the \modelName{}-distilled student seemingly features enhanced control and editability compared to the teacher model due to its one-step nature. Additionally, we present further results of latent space interpolation in \cref{sec:additional_qualitative}. In this process, the student model, distilled from the Stable Diffusion teacher, demonstrates a seamless transition in its image generation.

%% file: sec/conclusion.tex
\section{Conclusion and Discussion}\label{sec:discussion_conclusion}
% In this work, we have introduced a new image-free distillation scheme, \textbf{\modelName}, which leverages insights from text-to-3D synthesis to speed up text-to-image generation. Empirically, the results presented in this paper demonstrate the effectiveness of our approach in making text-to-image generation more efficient and accessible. We hope our work will encourage broader use of text-to-image generation systems by substantially lowering the inference cost.

% \modelName{} inevitably produces lower quality samples compared to the teacher which utilizes multi-step sampling for inference. Another drawback is that the current design only focuses on single-step student model and cannot support few-step generation as did in Luo et al. \cite{luo2023latent} to improve image fidelity. In future work, we aim to investigate the feasibility of extending \modelName{} to a few-step generation so that we can trade computation for quality. Furthermore, we also find it compelling to explore \modelName-like training where only one teacher is required. Such an exploration could provide tremendous benefits in scenarios where the computational resource is limited. Finally, it is intriguing to see if techniques such as DreamBooth \cite{ruiz2022dreambooth}, ControlNet \cite{zhang2023adding} or InstructPix2Pix\cite{brooks2022instructpix2pix} can be integrated with \modelName{}, allowing instant generation of various applications.

\minisection{Conclusion.} In this work, we have introduced a new image-free distillation scheme, \textbf{\modelName}, which leverages insights from text-to-3D synthesis to speed up text-to-image generation. Empirically, the results presented in this paper demonstrate the effectiveness of our approach in making text-to-image generation more efficient and accessible. We hope our work will encourage broader use of text-to-image generation systems by greatly lowering the inference cost.

\minisection{Discussion.} \modelName{} inevitably produces lower quality samples compared to the teacher which utilizes multi-step sampling for inference. Another drawback is that the current design only focuses on single-step student model and cannot support few-step generation as did in Luo et al. \cite{luo2023latent} to improve image fidelity. In future work, we aim to investigate the feasibility of extending \modelName{} to a few-step generation so that we can trade computation for quality. Furthermore, we also find it compelling to explore \modelName-like training where only one teacher is required. Such an exploration could provide tremendous benefits in scenarios where the computational resource for training is limited. Finally, it is intriguing to see if techniques such as DreamBooth \cite{ruiz2022dreambooth}, ControlNet \cite{zhang2023adding} or InstructPix2Pix\cite{brooks2022instructpix2pix} can be integrated with \modelName{}, allowing instant generation of various applications.

% \modelName{} produce lower quality outputs than its teacher, which uses multi-step sampling for inference. This model is currently limited to single-step generation, unlike Luo et al. \cite{luo2023latent} that enhance image quality through few-step generation. Future efforts will explore extending \modelName{} to few-step generation to balance computation and quality. Additionally, we are interested in a \modelName-like approach that requires only one teacher, which could be advantageous in compute-constrained environments. We also plan to investigate integrating techniques like DreamBooth \cite{ruiz2022dreambooth}, ControlNet \cite{zhang2023adding} or InstructPix2Pix\cite{brooks2022instructpix2pix} with \modelName{}, enabling instant generation of various applications.

%% file: sec/X_suppl.tex
\clearpage
\setcounter{page}{1}
\maketitlesupplementary

\section{Additional Details}\label{sec:training_estimate}
\minisection{Training Cost.} When training our model on a single A100 40GB GPU, we can use a batch size of up to 16 and approximately $15,825$ training iterations per day. Specifically, we train the model with $65,000$ gradient updates. Therefore, distilling Stable Diffusion using \modelName{} requires ${65,000}/15,825 \approx {4.11}$ A100 GPU days. Additionally, even though the original BOOT requires $500,000$ iterations to be fully trained as reported in \cite{gu2023boot}, our re-implementation only needs about $100,000$ to converge, which takes approximately $100,000 / 18,000 \approx 5.56$ A100 GPU day with a batch size of 16 per GPU.

\minisection{Inference Speed.} Next, we measure the inference time of our method using a batch size of $1$ on an A100 40GB GPU. The inference time of one-step UNet to generate the latent from random noise is $25$ ms, while those of text encoder and VAE are $80$ ms and $5$ ms, respectively. Consequently, the total inference time of \modelName{} is $25 + 80 + 5 = 110$ ms. Furthermore, the speed of the text encoder and VAE in BOOT are the same as ours; however, there is a small overhead in the UNet ($30$ms) as additional layers are used. Therefore, the total inference time for BOOT is $115$ms. 

\minisection{Memory Requirement.} Furthermore, we evaluate SwiftBrush's memory consumption during both training (with a batch size of 16) and inference (with a batch size of 1) on an A100 GPU. Since our approach involves training only a student and an additional low-rank LoRA weight for our LoRA teacher, it maintains a minimal memory footprint.

\minisection{Compare with Others.} We also sum up inference speed, training time and memory requirements of existing methods and ours in \cref{tab:speed_cost}. In terms of training costs, Guided Distillation and InstaFlow \cite{liu2023instaflow} require a considerable amount of training time, limiting their practical use. Conversely, LCM converges quickly but produces poor-quality images with one-step inference. Our method, however, offers both efficiency and high-quality output.

\begin{table}[t]
\centering
\resizebox{\columnwidth}{!}{ 
\begin{tabular}{lcccc}
\toprule
\multirow{2.5}{*}{Method} & \multicolumn{2}{c}{Training}              & \multicolumn{2}{c}{Inference}                       \\
\cmidrule(lr){2-3}\cmidrule(lr){4-5}
& Time (GPU days) & Memory (GB) & Time (ms) & Memory (GB) \\ 
\midrule
Guided Distillation & 108.8$^\dag$ & - & - & - \\
LCM & 1.3$^\dag$ & 33.6$^*$ & 118$^*$ & 11.3$^*$ \\
InstaFlow & 199.2$^\dag$ & - & 116$^*$ & 12.3$^*$ \\
\midrule
BOOT$^\ddag$ & 5.6 & 30.2 & 115 & 8.3 \\
Ours & 4.1 & 26.4 & 110 & 8.2 \\
\bottomrule
\end{tabular}
}
\centering
\vspace{-2mm}
\caption{Comparison of our method against other works on inference and training time/memory. $^\dag$ means that we obtain the numbers from the corresponding papers. $^*$ means that we obtain the numbers using the official code of the corresponding papers. $^\ddag$ means that we re-implement the work and report the numbers. The units for inference speed, training time and memory usage are miliseconds, A100 GPU days and gigabytes, respectively.}
\label{tab:speed_cost}
\vspace{-4mm}
\end{table}

\section{Additional Visual Results}\label{sec:additional_qualitative}
\begin{figure*}[t]
    \small
    \begin{center}
        \includegraphics[width=\textwidth]{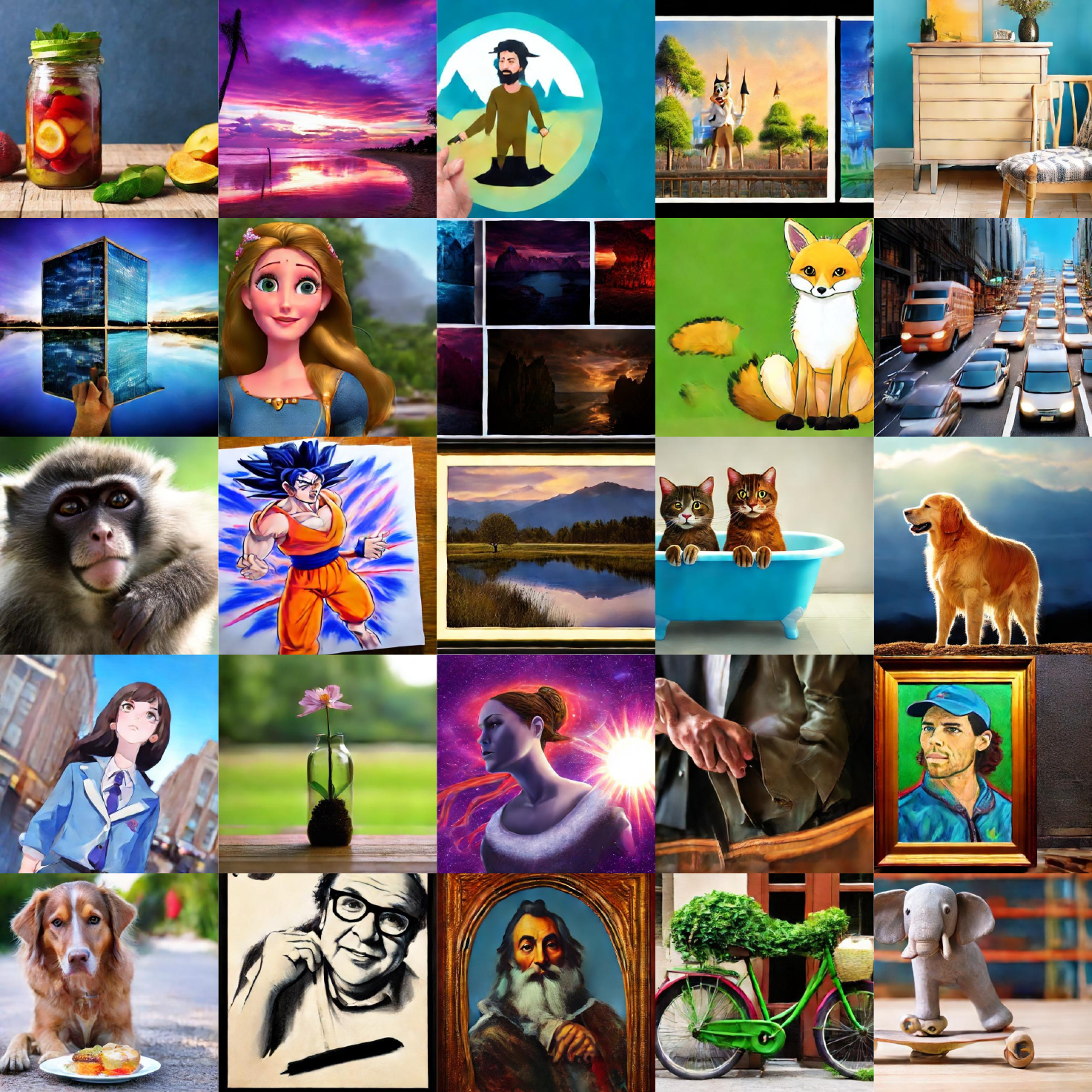}
    \end{center}
    \vskip -0.22in
    \caption{Uncurated samples from one-step \textbf{\modelName{}}.}
    \label{fig:supp_ours}
    \vskip -0.15in
\end{figure*}

\begin{figure*}[t]
    \small
    \begin{center}
        \includegraphics[width=\textwidth]{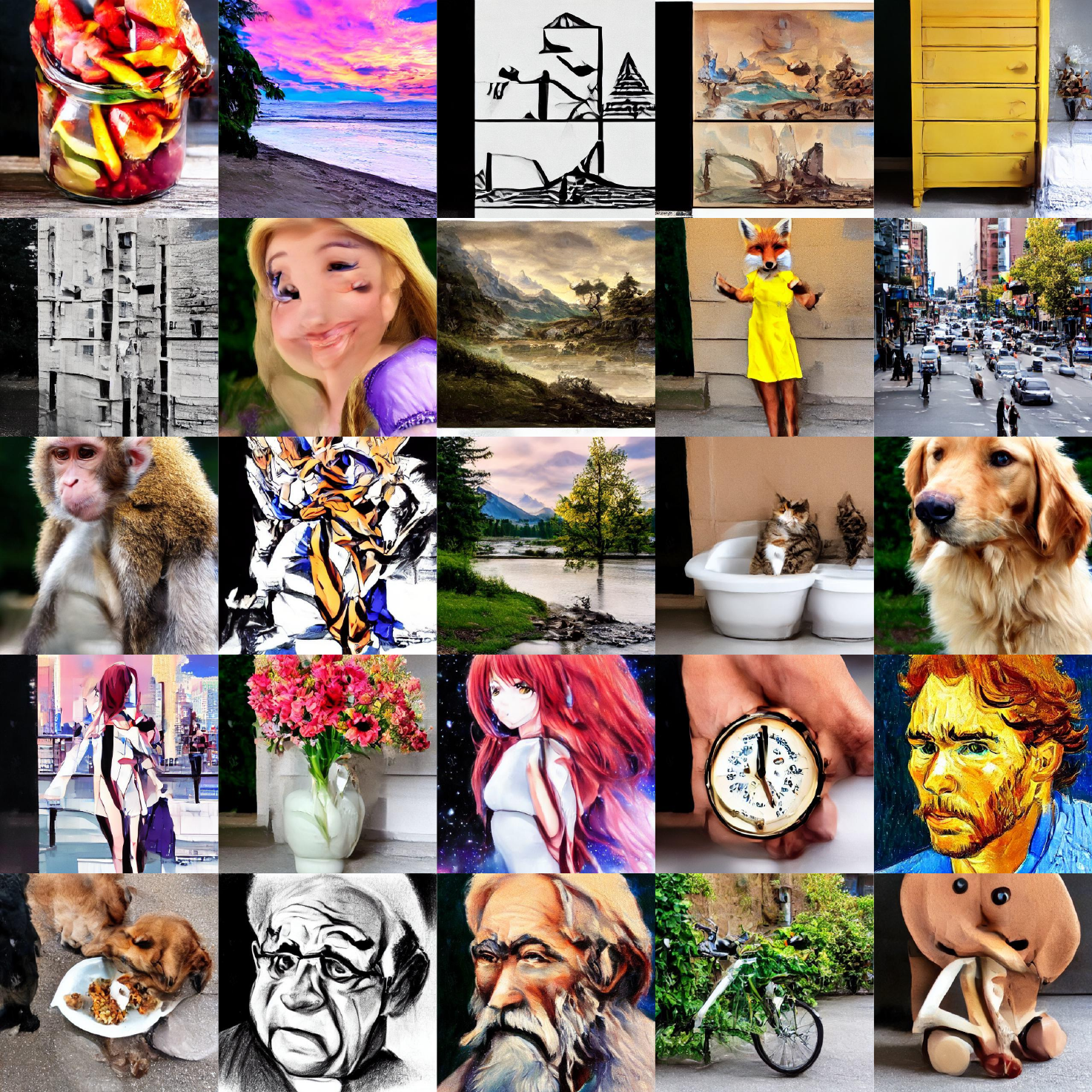}
    \end{center}
    \vskip -0.22in
    \caption{Uncurated samples from one-step \textbf{InstaFlow}. We use their provided pretrained model to generate.}
    \label{fig:supp_instaflow}
    \vskip -0.15in
\end{figure*}

\begin{figure*}[t]
    \small
    \begin{center}
        \includegraphics[width=\textwidth]{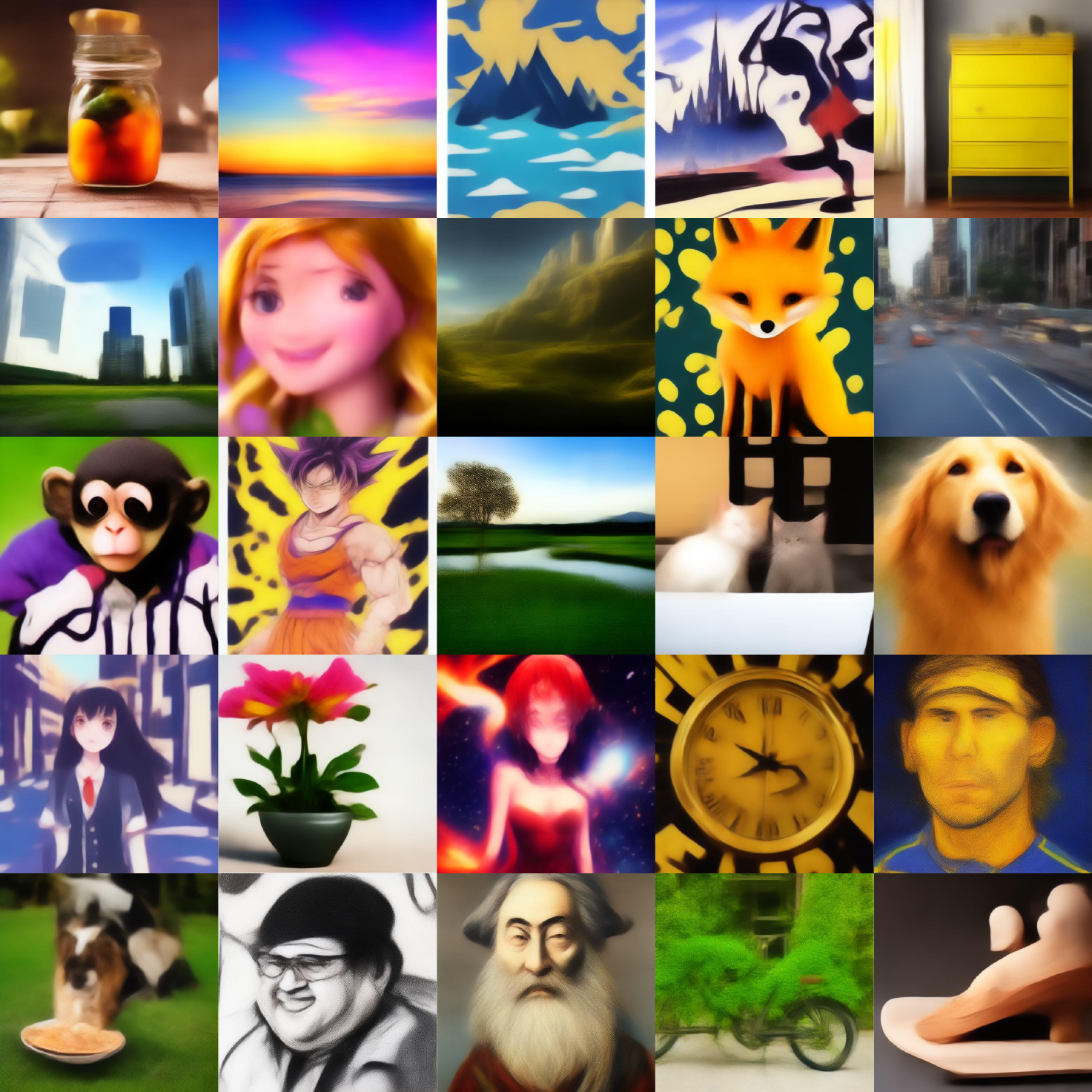}
    \end{center}
    \vskip -0.22in
    \caption{Uncurated samples from one-step \textbf{BOOT}. We re-implement and generate the images by ourselves.}
    \label{fig:supp_boot}
    \vskip -0.15in
\end{figure*}

\begin{figure*}[t]
    \small
    \begin{center}
        \includegraphics[width=\textwidth]{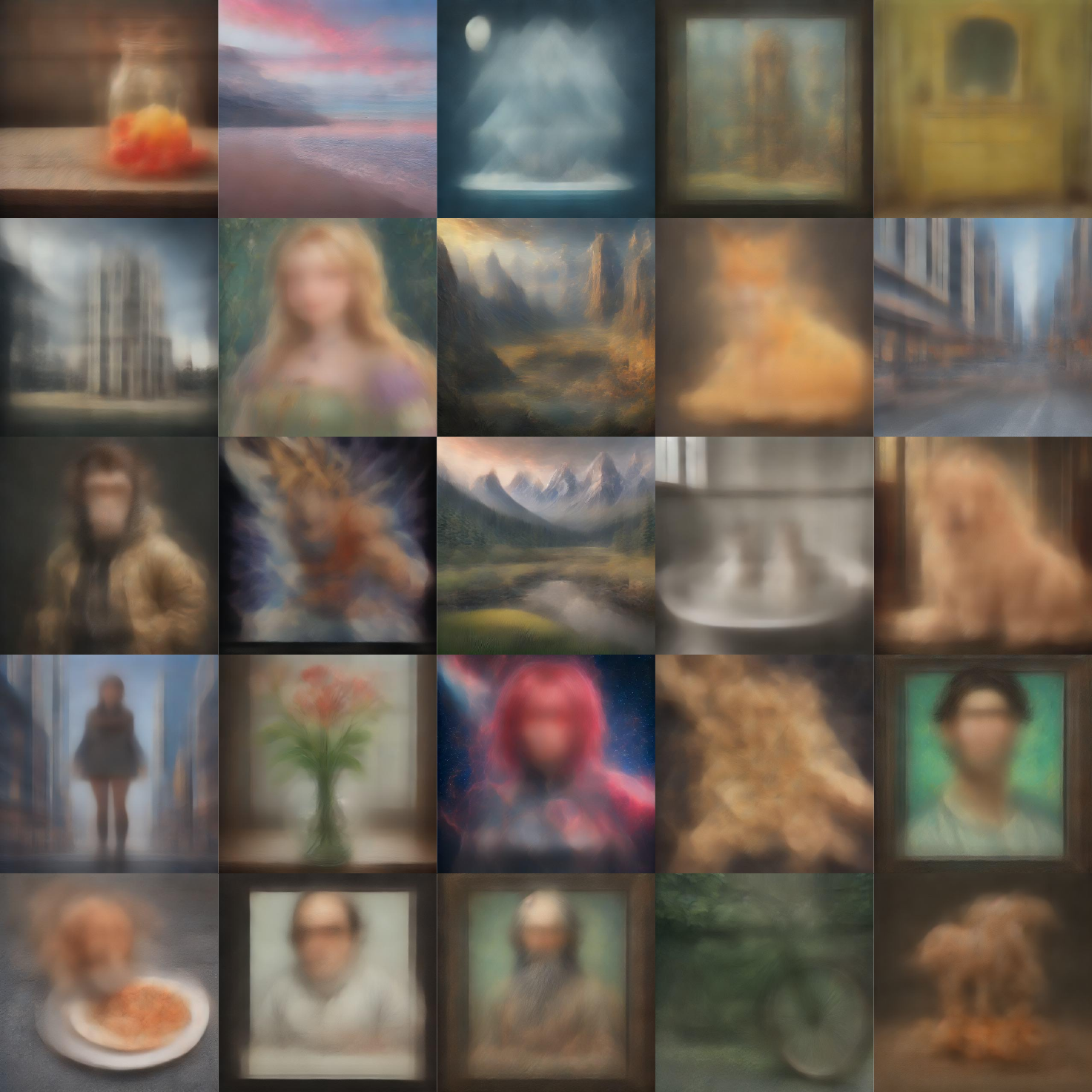}
    \end{center}
    \vskip -0.22in
    \caption{Uncurated samples from one-step \textbf{LCM}. We use their provided pretrained model to generate.}
    \label{fig:supp_lcm}
    \vskip -0.15in
\end{figure*}

\minisection{Uncurated Samples.} We display uncurated images conditioned on 25 random text prompts\footnote{\href{https://vinairesearch.github.io/SwiftBrush/assets/prompts.txt}{https://vinairesearch.github.io/SwiftBrush/assets/prompts.txt}}. The images generated by various models are presented in \cref{fig:supp_ours}, \labelcref{fig:supp_instaflow}, \labelcref{fig:supp_boot} and \labelcref{fig:supp_lcm} for a better visual comparison between our model and others.

\begin{figure*}[t]
    \small
    \begin{center}
        \includegraphics[width=\textwidth]{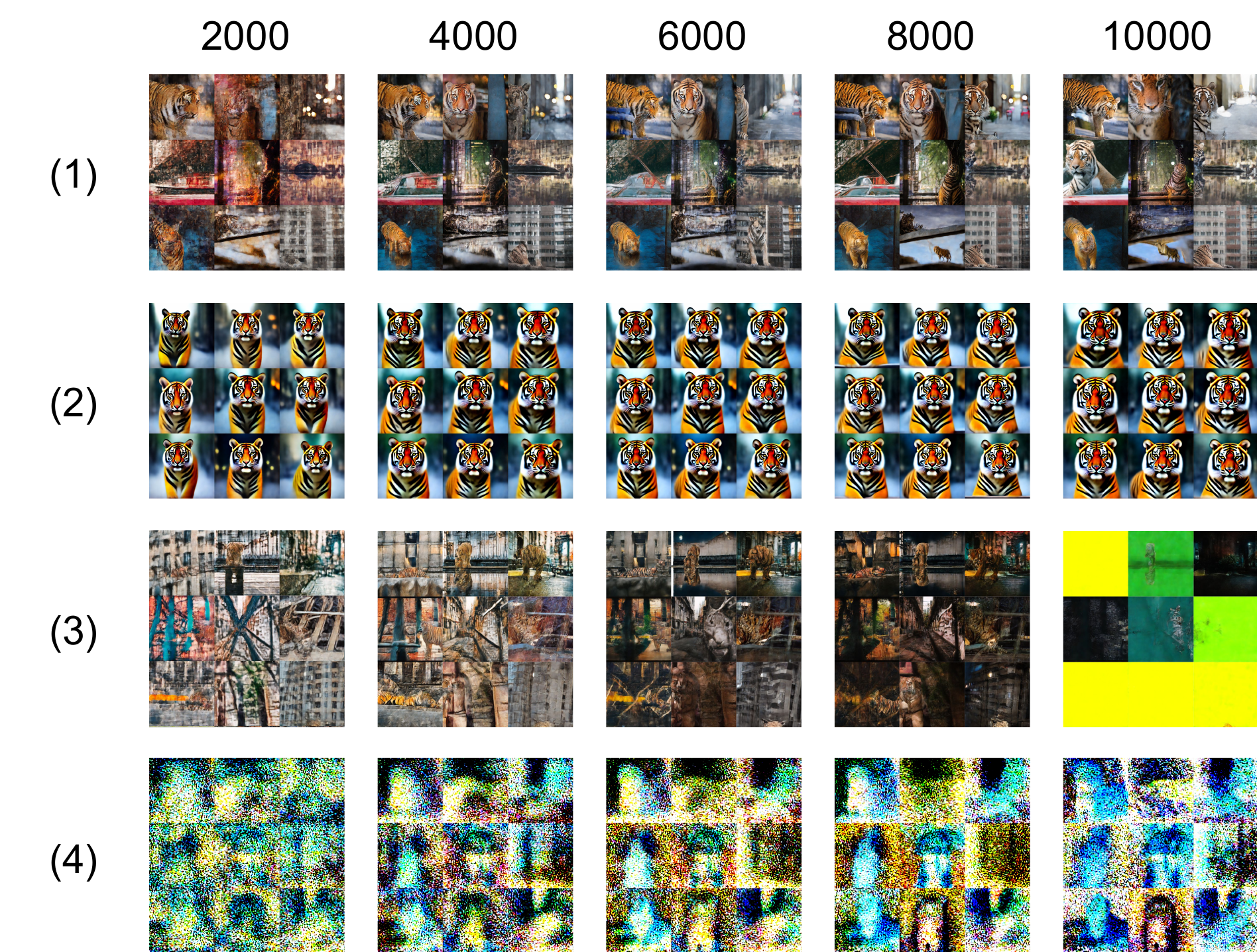}
    \end{center}
    \vskip -0.22in
    \caption{Visual results of the ablative study, where the all images are generated with the input prompt \textit{``An DSLR photo of a tiger in the city``}. Images at same column are generated at the same iteration of training, whereas those at same row are generated using the same configuration. Here, (1) means ``Full'', (2) means ``SDS'', (3) means ``Small-Rank LoRA'' and (4) means ``No Parameterization''.}
    \label{fig:supp_ablate}
    \vskip -0.15in
\end{figure*}

\minisection{Additional Visuals of Analysis.} To further back up our claim in \cref{sec:analysis}, we provide some uncurated samples using the text prompt \textit{``An DSLR photo of a tiger in the city``}. The images generated by four configurations at every 2000 iterations are presented in \cref{fig:supp_ablate} for a better visual comparison of each component's importance in our proposed method across different training stages.

\begin{figure*}[t]
    \small
    \begin{center}
        \includegraphics[width=\textwidth]{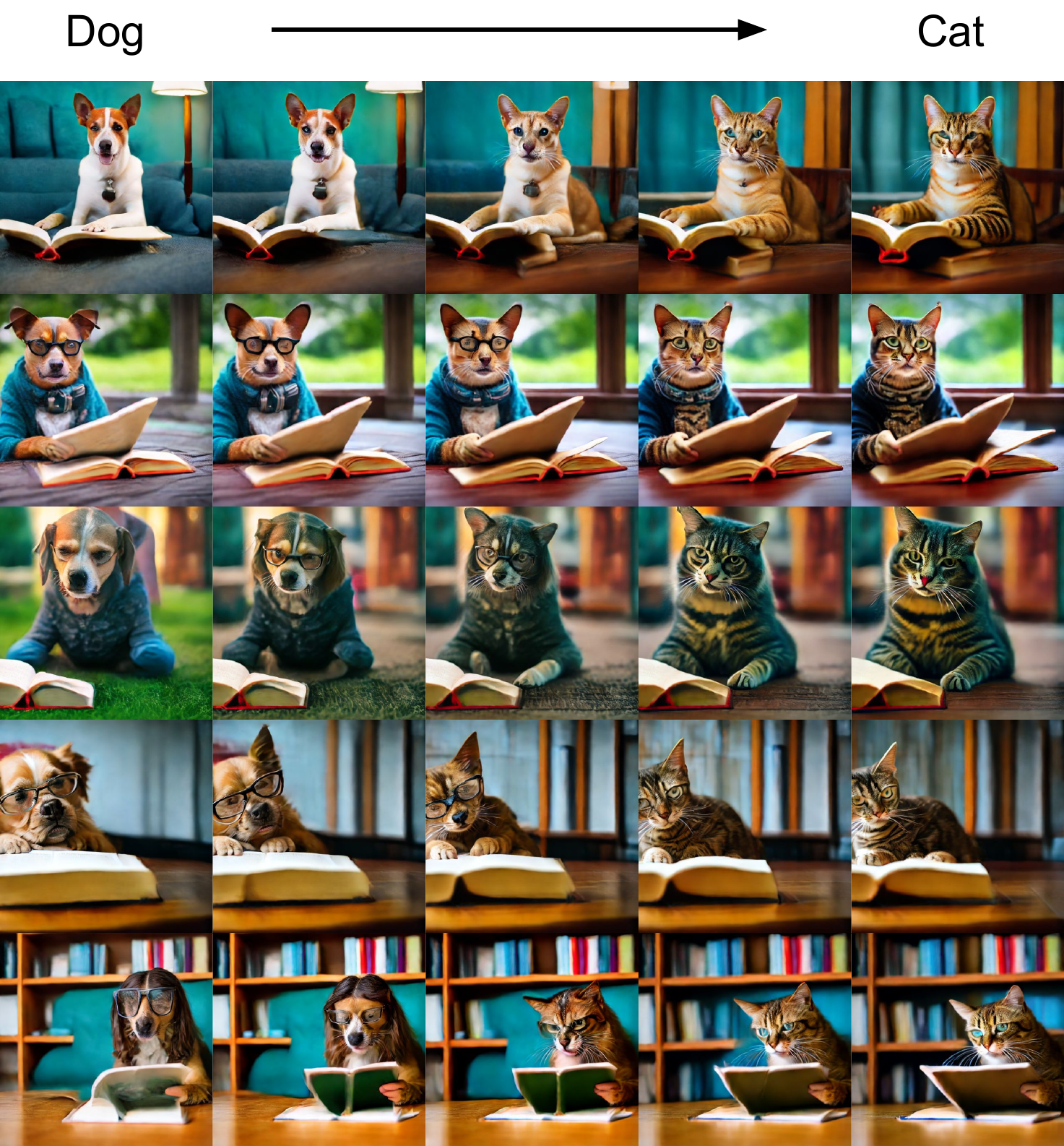}
    \end{center}
    \vskip -0.22in
    \caption{Results of interpolating the input prompt. The prompts
used here are selected from a standard template: \textit{``A DSLR photo of a \{animal\} reading a book''}. Here, ‘animal’ is dog or cat and we interpolate the text embedding using linear interpolation (Lerp). Same noise input $z$ is used for images at each row.}
    \label{fig:supp_interpolate_text}
    \vskip -0.15in
\end{figure*}

\begin{figure*}[t]
    \small
    \begin{center}
        \includegraphics[width=\textwidth]{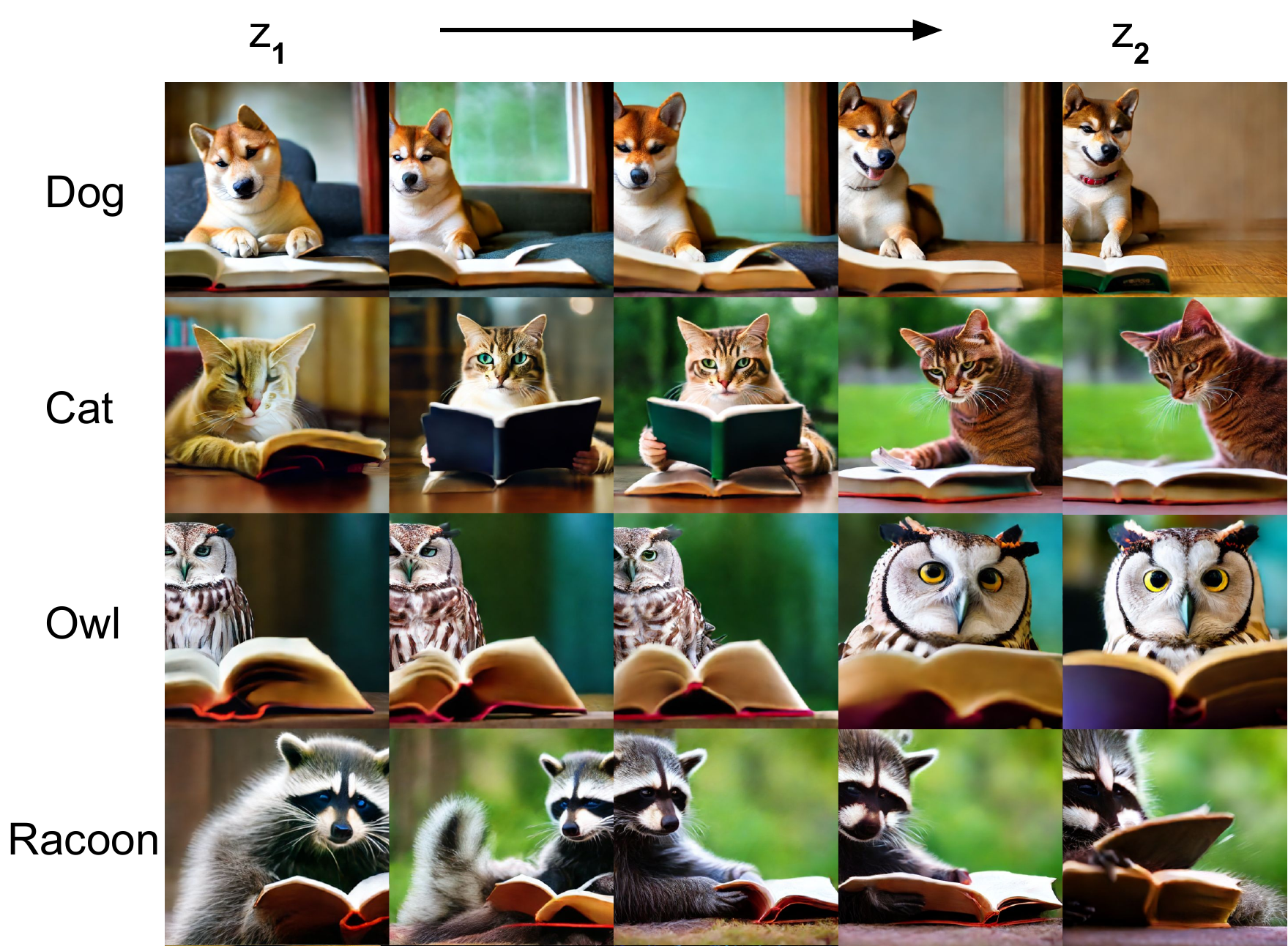}
    \end{center}
    \vskip -0.22in
    \caption{Results of interpolating the noise input. The prompts
used here are selected from a standard template: \textit{``A DSLR photo of a \{animal\} reading a book''}. Here, ‘animal’ is dog, cat, owl or racoon and we interpolate the noise input using spherical linear interpolation (Slerp). Same text input $y$ is used for images at each row.}
    \label{fig:supp_interpolate_noise}
    \vskip -0.15in
\end{figure*}

\minisection{Additional Visuals of Interpolation.} Apart from investigating the role of each component in \modelName{}, we also explore its properties when interpolating text prompt condition $y$ (\cref{fig:supp_interpolate_text}) or noise input $z$ (\cref{fig:supp_interpolate_noise}). It is evident that our model generates a seamless transition in the output, showcasing excellent editability and controllability.

\section{Additional Quantitative Results}\label{sec:additional_quantitative}
We run SwiftBrush on CIFAR-10 $32\times32$ and class-conditional ImageNet $64\times64$ benchmarks using EDM teacher models and compare it with Progressive Distillation (PD) and Consistency Distillation (CD), either using L2 or LPIPS loss, in \cref{tab:supp_fid}. SwiftBrush is on par with CD-LPIPS and it significantly outperforms the others.

\begin{table}[b]
\centering
\footnotesize

\begin{tabular}{l|c|c|c}
\hline
Method & Steps & CIFAR-10 $32\times32$ & ImageNet $64\times64$ \\
\hline
PD & 1 & 8.34 & 15.39 \\
CD - L2 & 1 & 7.90 & 12.10 \\
CD - LPIPS & 1 & \textbf{3.55} & \underline{6.20} \\
Ours & 1 & \underline{4.46} & \textbf{5.85} \\
\hline
EDM & 35 & 1.97 & - \\
EDM & 79 & - & 2.44 \\
\hline
\end{tabular}
\vspace{-2mm}
\caption{Comparison of our method against other works in FID on CIFAR-10 ($32\times32$) and ImageNet ($64\times64$) benchmarks.}
\label{tab:supp_fid}
\vspace{-2mm}
\end{table}